\documentclass[conference]{IEEEtran}
\IEEEoverridecommandlockouts
\usepackage{cite}
\usepackage{amsmath,amssymb,amsfonts,mathtools}
\usepackage[caption=false,font=footnotesize,labelfont=sf,textfont=sf]{subfig}
\usepackage{float}
\usepackage{algorithmic}
\usepackage{graphicx}
\usepackage{textcomp}
\usepackage{xcolor}

\def\BibTeX{{\rm B\kern-.05em{\sc i\kern-.025em b}\kern-.08em
    T\kern-.1667em\lower.7ex\hbox{E}\kern-.125emX}}
\begin{document}

\title{Hierarchical Conditional Variational Autoencoder \\Based Acoustic Anomaly Detection

}

\author{\IEEEauthorblockN{Harsh Purohit}
\IEEEauthorblockA{ 
\textit{R\&D Group, Hitachi Ltd.}\\
Tokyo, Japan \\
harsh\_pramodbhai.purohit.yf\\@hitachi.com}
\and
\IEEEauthorblockN{Takashi Endo}
\IEEEauthorblockA{
\textit{R\&D Group, Hitachi Ltd.}\\
Tokyo, Japan \\
takashi.endo.qf\\@hitachi.com}
\and
\IEEEauthorblockN{Masaaki Yamamoto }
\IEEEauthorblockA{
\textit{R\&D Group, Hitachi Ltd.}\\
Tokyo, Japan \\
masaaki.yamamoto.af\\@hitachi.com}
\and
\IEEEauthorblockN{Yohei Kawaguchi}
\IEEEauthorblockA{
\textit{R\&D Group, Hitachi Ltd.}\\
Tokyo, Japan \\
yohei.kawaguchi.xk\\@hitachi.com}

}

\maketitle

\begin{abstract}
	This paper aims to develop an acoustic signal-based unsupervised anomaly detection method for automatic machine monitoring. Existing approaches such as deep autoencoder (DAE), variational autoencoder (VAE), conditional variational autoencoder (CVAE)  etc. have limited representation capabilities in the latent space and, hence, poor anomaly detection performance. Different models have to be trained for each different kind of machines to accurately perform the anomaly detection task. To solve this issue, we propose a new method named as hierarchical conditional variational autoencoder (HCVAE). This method utilizes available taxonomic hierarchical knowledge about industrial facility to refine the latent space representation. This knowledge helps model to improve the anomaly detection performance as well. We demonstrated the generalization capability of a \textit{single} HCVAE model for different types of machines by using appropriate conditions. Additionally, to show the practicability of the proposed approach, (i) we evaluated HCVAE model on different domain and (ii) we checked the effect of partial hierarchical knowledge. Our results show that HCVAE method validates both of these points, and it outperforms the baseline system on anomaly detection task by utmost 15 \% on the AUC score metric.
\end{abstract}

\begin{IEEEkeywords}
Hierarchical variational autoencoder, Acoustic anomaly detection,  Unsupervised anomalous sound detection.
\end{IEEEkeywords}

\section{Introduction}
  Recently, Internet of Things (IoT) and data driven techniques have been transforming the manufacturing industry and different approaches have been undertaken for automatic factory maintenance. These approaches are mainly based on different sensor data parameters. For example, temperature based \cite{lodewijks2016application}, pressure based \cite{salikhov2019diagnosis} and image based methods \cite{janssens2015thermal}. Alternatively, due to low installation cost and contact-less sensors, Anomalous Sound Detection (ASD) is currently a focused field for audio surveillance \cite{foggia2015audio}, predictive maintenance\cite{yamashita2006inspection}, and automatic machine inspection \cite{koizumi2018unsupervised,Koizumi2020,kawaguchi2019anomaly,purohit2020deep}. In this study, we investigated ASD for industrial machines by focusing on its operating sound signals.

 Deep autoencoders (DAE) are particularly adopted for unsupervised anomaly detection \cite{Koizumi2020}. When trained on normal data, testing on unseen faulty samples tends to yield suboptimal representation, indicating that a sample is likely generated by an abnormal operating condition. Moreover, variatonal autoencoder (VAE) \cite{kingma2013auto} based technique is also useful to accurately represent the normal machine sound captured through a microphone. We observed that anomaly detection task is dependent on the latent space representation in VAE. For specific sub-cases like anomaly detection, VAE architecture may not be adequate and it must be tweaked appropriately \cite{wang2019self}. This motivates us to include the available knowledge into the training procedure of VAE and enhance the representation. As a result of that, we propose a new model called hierarchical conditional variational autoencoder (HCVAE) which also helps to improve the accuracy of anomaly detection task in machine sounds.


	    We evaluated the proposed method on two scenarios to show the practicability. (i) We checked the domain adaptation capabilities of the model. For this, we first trained the model on domain-1 and then utilized those weights for adapting them to domain-2. Domain-1 and domain-2 have similar kind of machines but have difference in terms of manufacturers. (ii) We evaluated the model on partial taxonomic knowledge. In real life scenario, it is possible that we only have partial information instead of full hierarchical knowledge in our data. In that case also, we show that the anomaly detection performance can be improved using HCVAE model.
	    
	    The rest of this paper is organized as follows. Section II briefly introduces the background of variational autoencoders. Section III describes the proposed hierarchical conditional variational autoencoder method, and the details of the implementation are discussed in Section IV. After reporting results and discussion in Section V, we conclude this paper in Section VI. 


\section{Background of variational autoencoders}

	Variational autoencoders are generative models introduced in \cite{kingma2013auto} by Kingma and Welling. VAE allow us to effectively compress information within the data through dimensionality reduction occurring in the latent space. VAE consists of an encoder-decoder pair of neural networks, with a stochastic latent layer in the middle. The encoder which we express here with $p_{\theta}(z|x)$ processes the input $x$ and produces the parameters of a Gaussian distribution represented by $z$ in the latent layer. The decoder which we denote by $p_{\theta}(x'|z)$ uses that Gaussian distribution $z$ in latent space as input to produce the parameters to an approximation of the probability distribution describing the original data. A prime advantage of this approach is that VAE, unlike regular autoencoders, map values to the latent space by retaining relational information between them. Such model can be used for density estimation and subsequently for anomaly detection task.


    
\vskip 0.07in	
\textbf{The objective function in VAE:}
\vskip 0.1in
 The choice of loss function during any optimization or neural network problem is critical. Since the decoder goes from a smaller to a larger space, information is lost. We measure this loss by the reconstruction loss which is the log-likelihood of $p_{\theta}(x'|z)$ which will give us a loss function to minimize in training. This shows also how effective the encoder was in compressing information within $z$ from the original data $x$. However, to create a proper loss function, one requires knowledge of $p_{\theta}(x)$ which is intractable \cite{kingma2013auto}. To counter this difficulty, we consider a lower bound of the likelihood and optimize that instead.

   Our aim here is to model the data, hence we want to find the distribution  $p_{\theta}(x)$. Using the law of probability, we could find it in relation with $z$ as follows:
   
   	\begin{align}
     p_{\theta}(x) = \int p_{\theta}(x | z) p_{\theta}(z) dz.
	\end{align}
   
 Using the KL \footnote{KL divergence metric is useful to find the difference between two distribution P and Q and it is defined as $D_{KL}(P||Q)= \mathbb{E}_{x\in P}\Big[log \frac{P(x)}{Q(x)}\Big]$.}  divergence, we can express $\log p_{\theta}(x)$ as,
	
	\begin{align}
	\begin{split}
     \log p_{\theta}(x) =  \mathbb{E}_{z\in q_{\phi}(z|x)}[\log  p_{\theta}(x|z)] - D_{KL} 	\\ \big( q_{\phi}(z|x)||p_{\theta}(z) \big) +D_{KL} \big( q_{\phi}(z|x)||p_{\theta}(z|x) \big).
     \end{split}
     \label{eq2}
	\end{align}

	The first term above is possible to compute directly from decoder and sampling with the parameterization trick \cite{kingma2013auto}. The second term is a KL-divergence between a general and a normal Gaussian distribution which has a closed form solution. Finally, the last KL-divergence form can be shown greater than 0 \cite{kingma2013auto}. Therefore, lower bound of equation \ref{eq2} can be defined as, 
	
	\begin{align}
	\begin{split}
	      L (\theta, \phi ; x) =  \mathbb{E}_{z\in q_{\phi}(z|x)}[\log  p_{\theta}(x|z)] - D_{KL} \\ \big( q_{\phi}(z|x)||p_{\theta}(z) \big).
	\end{split}
    \label{eq3}
	\end{align}
	
	Thus, our task is to minimize the above with respect to the parameters $\theta$ and $\phi$ as,
	
	\begin{align}
     \theta^*, \phi^*  = \operatorname*{argmin}_{\theta,\phi} L (\theta, \phi ; x).
     \label{eq4}
	\end{align}

\section{Hierarchical conditional variational autoencoder}
	
		\begin{figure}
	\begin{center}
		\includegraphics[height=3.5cm,width=7.5 cm]{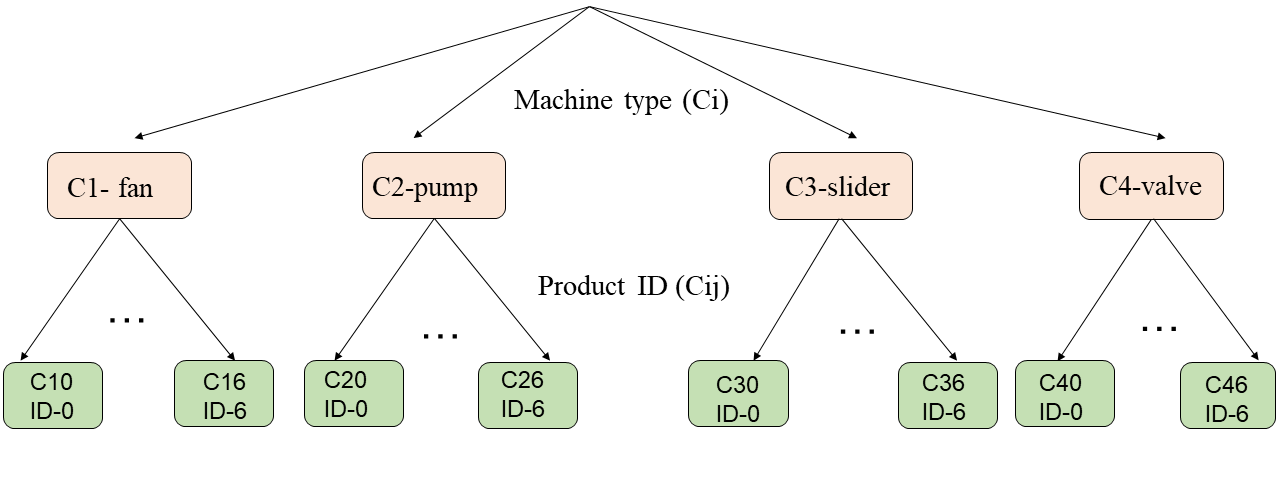}\vspace{-0.3cm}
		\caption{Hierarchical structure in the MIMII dataset \cite{Purohit2019}} 
		\label{hiera}
	\end{center}
    \end{figure}
	
	Although VAE offers a more general and principal way for density estimation, it could suffer from  intrinsic limitations when our goal is anomaly detection. Hence, we propose to use taxonomic knowledge information which is already available in the data. By including the associated known or observable variables, we can efficiently train our model.

\begin{figure*}
\centering
\subfloat[Fan IDs 0,2,4,6 without condition]{\includegraphics[width=2.15in]{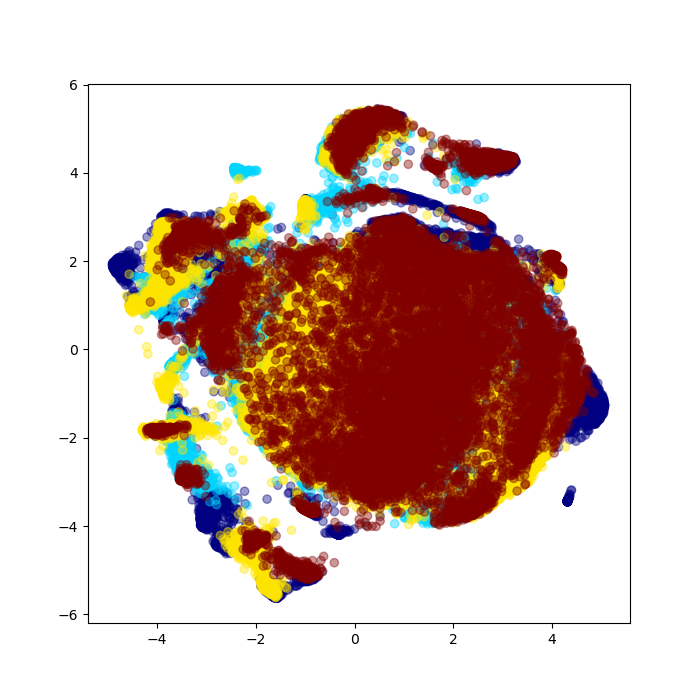}}
\hfil
\subfloat[Fan IDs 0,2,4,6 with condition $C_{ij}$]{\includegraphics[width=2.15in]{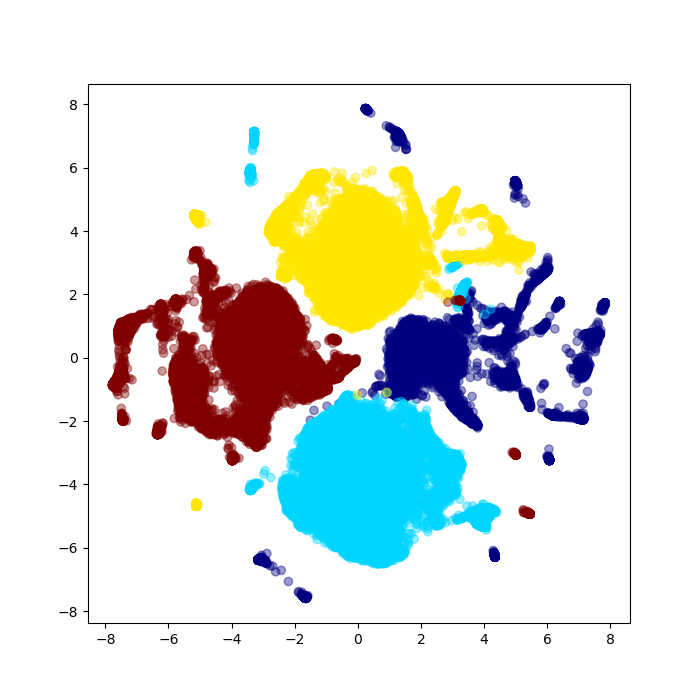}}

\caption{Illustration of AUC for different machines in evaluation set.}
 \label{latent} 
\end{figure*}

	We make our discussion more concrete by an example. As shown in Fig \ref{hiera}, MIMII dataset \cite{Purohit2019} contains two level hierarchies in its structure. The first level represents machine type, i.e. fan, pump, slider, valve and second level has different machine IDs for each machine type. These inherent taxonomic knowledge with level one $C_i$ (machine type) and level two $C_{ij}$ (model ID) can be utilized for additional information to our model. Due to these additional inputs, our model is capable of learning the latent representation very accurately. For example, Fig.
	\ref{latent} shows latent representation (in projected 2D space) before and after the inclusion of $C_i$ condition.

    
    
\begin{figure} [H]
	\begin{center}
		\includegraphics[height=3.5cm,width=7.5 cm]{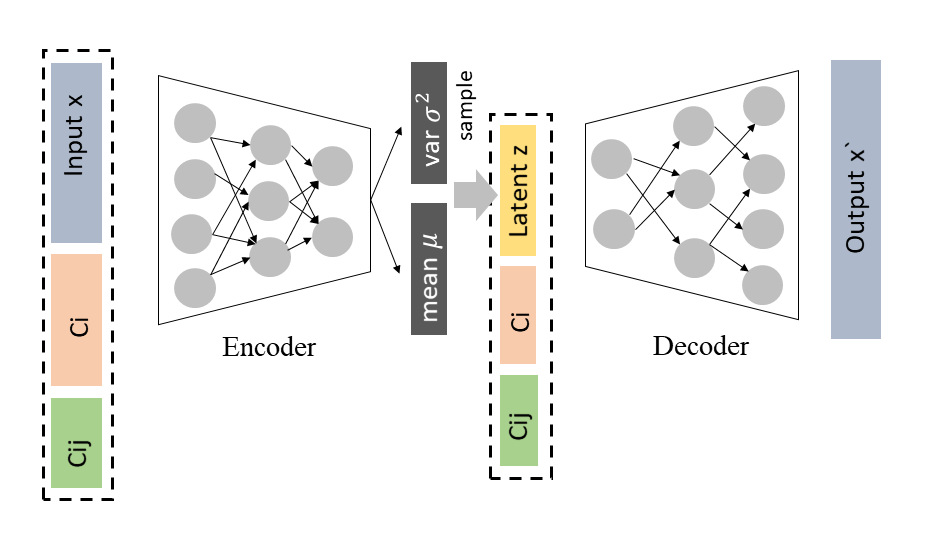}\vspace{-0.3cm}
		\caption{Diagram of an HCVAE structure.} 
		\label{hcvae}
	\end{center}
    \end{figure}

	   The proposed hierarchical conditional variational autoencoder (HCVAE) is an extension of VAE. In VAE, we have no control on the latent  space representation. Hence, we propose HCVAE which models latent variable based on available information as input conditions. Fig. \ref{hcvae} represents that condition $C_i$ and $C_{ij}$ are included with input vector $x$. Here, $x$ is made up of a collection of frames from log-mel spectrogram. Note that the same conditions are also merged with latent vector $z$ which is input to decoder network. In this case, $C_i$ and $C_{ij}$ are categorically distributed, or in other words, it takes form as a one-hot vector. However, one can adopt hash function if fix size vector is required in the application.

We understand that because each data point $x$ under specific condition has its own distribution, it is effective to represent it in latent space. The latent space will retain relational information of features and subsequently helps in better density estimation of complicated data. Due to this enhancement in the latent space representation, anomaly performance will also be improved.	
\vskip 0.07in	
\textbf{The objective function in HCVAE:}
\vskip 0.1in
  VAEs are defined to estimate the unknown probability distribution $p_{\theta}(x)$ for the given $x$ using a Gaussian in latent space. Instead, we proposed to use hierarchical information with already available in the data.

    The original VAE model has two parts: the encoder $q_{\phi}(z|x)$ and the decoder $p_{\theta}(x|z)$. We observe that encoder models the latent variable $z$ directly based on $x$ and decoder part only models $x$ directly based on the latent variable $z$. We propose to improve VAE by conditioning the encoder and decoder to another information $C$. In our case, $C$ can be defined as $C = [C_i, C_{ij}]$. Where $C_i$ and $C_{ij}$ are level one and level two information as shown in Fig. \ref{hiera}. The encoder can now be conditioned to two variables $x$ and $C$: $q_{\phi}(z|x, C)$. The same with the decoder, it can be conditioned with two variables $z$ and $C$: $ p_{\theta}(x|z, C)$

 We conditioned all the distribution with variable $C$. Therefore, loss function can be described as follows:
 	\begin{align}
 	\begin{split}
 	   L (\theta, \phi ; x) =  \mathbb{E}_{z\in q_{\phi}(z|x)}[\log  p_{\theta}(x|z,C)] -  \beta D_{KL}\\ \big( q_{\phi}(z|x,C)||p_{\theta}(z|C) \big).
 	\end{split}
    \label{eq5}
	\end{align}
 Where $\beta$ is a scalar weight for KL-divergence term. Our task is to minimize the above with respect to the parameters $\theta$ and $\phi$ as equation \ref{eq5}.

	
	
	

	\section{Experiments}

In this paper, we use MIMII (Malfunctioning Industrial Machine Investigation and Inspection) dataset \cite{Purohit2019} to evaluate our proposed method.  MIMII dataset was released in 2019, and it was used in task-2 in DCASE 2020 challenge\cite{Koizumi2020}. This challenge task is known as the Unsupervised Detection of Anomalous Sounds for Machine Condition Monitoring.  The goal of this task is to have a quick detection of anomalous condition by observing the machine sound. The DCASE 2020 task-2 used two datasets: the MIMII dataset
\cite{Purohit2019} and the ToyADMOS dataset \cite{koizumi2019toyadmos}. However, in this work, we chose to focus on the MIMII dataset, since it consists of the sound data of real machines with background environment conditions. 

\subsection{Dataset attributes}
The MIMII dataset in task 2 of the DCASE 2020 challenge consists of audio data that had been recorded from four types of machines. Each machine type had seven  different machine IDs. Under the running condition, the 10-second sound of the machine was recorded as 16-bit audio signals sampled at 16 kHz in a reverberant environment. 

\subsubsection{Dataset formation}
The MIMII dataset as used in task 2 of DCASE 2020 challenge was presented in the form of 3 sets:
\begin{itemize}
\item Development dataset:
    \begin{itemize}
        \item Training set: as training data, only normal data, labeled, intended to be used to train models. 
        \item Test set: as test data, normal and anomaly data, labeled, intended to be used to test the trained models.
    \end{itemize}
\item Evaluation dataset: as test data, normal and anomaly data, not labeled.
\item Additional training dataset:  as training data, can be used for training, however this set was open later in the challenge.

\end{itemize}

In the current work, we are focusing on the development data and evaluation data. In the development dataset, we have the data of the 4 types of machines: valve, pump, fan and slide-rail (slider). And for every machine type, we have 4 machine IDs: 00, 02, 04 and 06. In evaluation dataset, the samples from same machine types' but different IDs (01, 03, 05) are available. The number of test samples for each machine ID is around 400, none of which have a label (i.e., normal or anomaly). Additionally, around 1,000 normal samples for each machine type and machine ID (1, 3, 5) used in the evaluation dataset are also used for domain adaptation test. Then, in total, we are training and testing our predictive scheme on 28 different machines. Therefore, we can verify ability to recognize the anomalous sounds by having normal data only as reference and over different machine types. Accordingly, we can verify the generalization and scalability of our proposed approach.

\subsection{Pre-processing}

Before sent into the neural network model, each audio needs to be pre-processed. A sliding window with 1024 frame size and 512 hop size is used to convert each audio into a spectrogram. Then a mel filter bank with 128 frequency bins is applied to the spectrogram to obtain the mel-spectrogram. Finally, the mel-spectrogram is converted to a logarithmic scale. Then, five frames were successively concatenated to make an input feature vector, with four overlapping frames. 

\begin{figure*}
\centering
\subfloat[Results of fans.]{\includegraphics[width=2.25in]{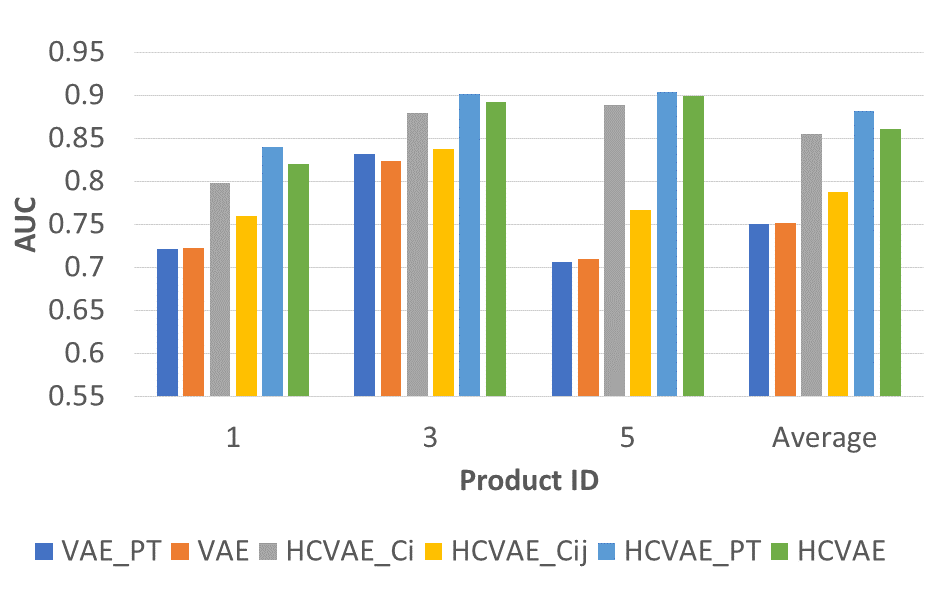}%
\label{fig_first_case}}
\subfloat[Results of pumps.]{\includegraphics[width=2.25in]{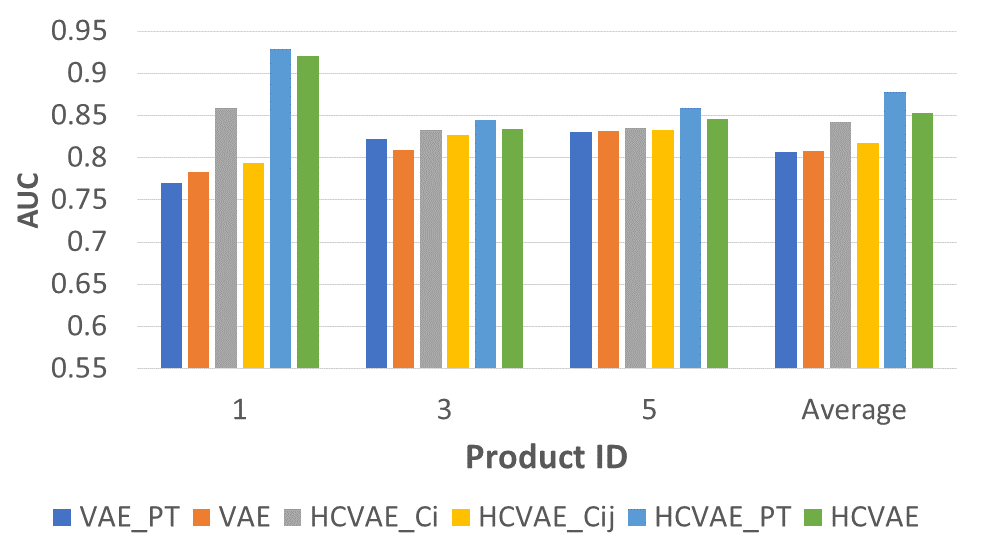}%
\label{fig_second_case}}
\hfil
\subfloat[Results of slide-rails.]{\includegraphics[width=2.25in]{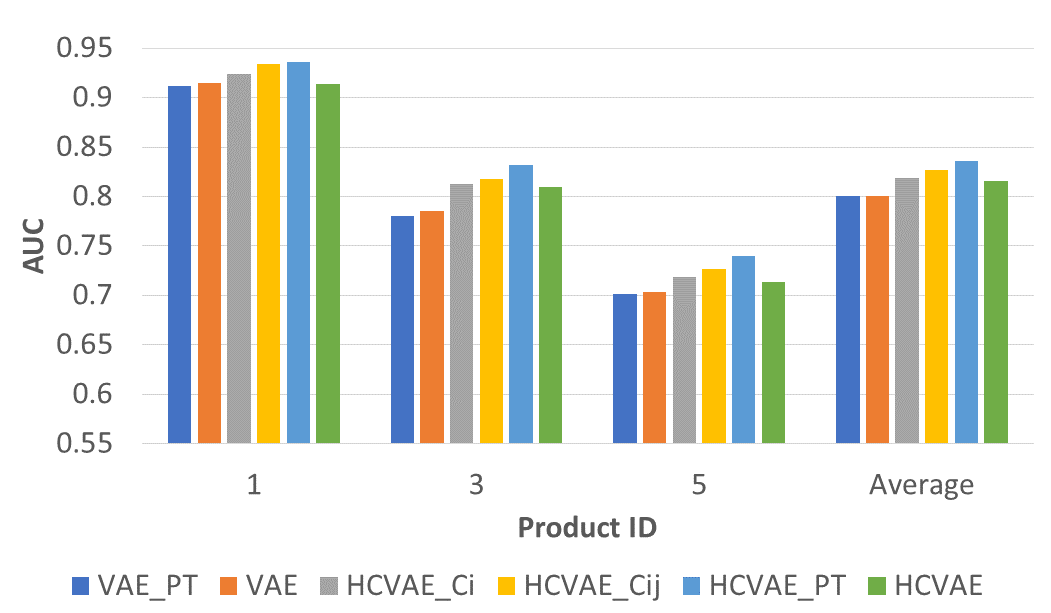}%
\label{fig_first_case}}
\subfloat[Results of valves.]{\includegraphics[width=2.25in]{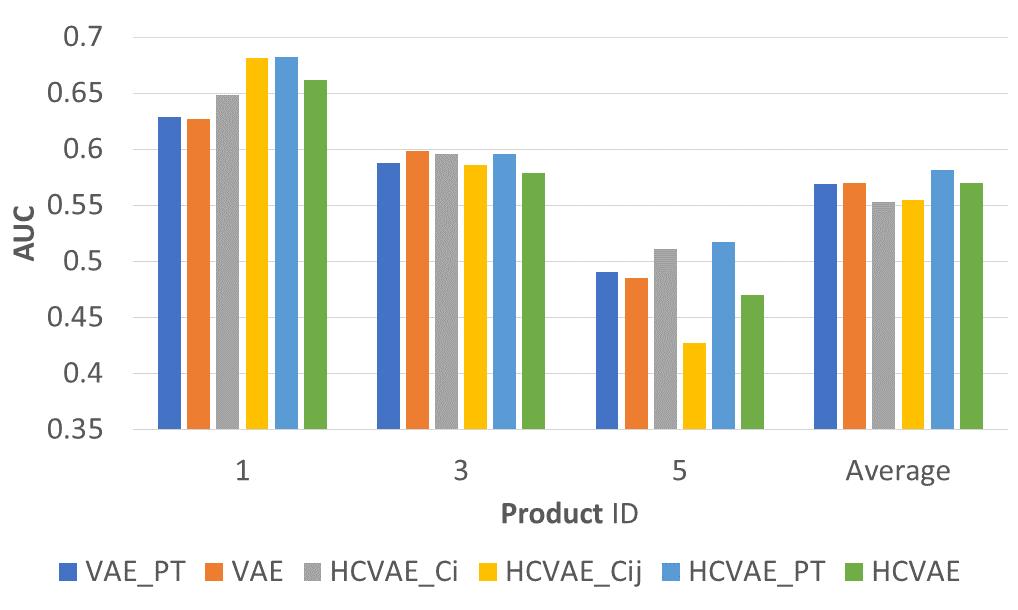}%
\label{fig_second_case}}
\caption{Illustration of AUC for different machines in evaluation set.}
\label{fig_si}
\end{figure*}

\subsection{Experiment Setting}

We implement the proposed model using Keras (Tensorflow as backend) on a PC having 32 GB RAM and Core i7 processor with single NVIDIA GeForce RTX 2080 super GPU card with. Adam optimizer \cite{kingma2014adam} with an initial learning rate of 0.001 is used for stable convergence. Moreover, we  evaluated it on the following two criteria: 
	
	\begin{itemize}
	    \item First, we checked the domain adaptation capability of HCVAE pre-trained model. In this test, we first train the HCVAE model on the domain-1 (i.e. for IDs 00, 02, 04, 06). Then we used the same weights from this model and trained the model on domain-2 (i.e. IDs 01, 03, 05) and test.
        \item Second, we trained the HCVAE model on partial taxonomic knowledge instead of full hierarchical information. In other words, we either give level one condition $C_i$ or level two condition $C_{ij}$ during training and examine the effect on anomaly detection performance. 
	\end{itemize}

\subsection{Evaluation Metric}

Background on the used metrics to evaluate the performance of our, proposed approach we used AUC metric.
The AUC refers to the area under the receiver operation characteristic (ROC) \cite{fawcett2006introduction} curve (AUC). The ROC graph has been known to be used to evaluate the performance of a two class classifier that maps instance the prediction to true positives and false positives. An instance is positive (abnormal) if the score is higher than the threshold. Else, if the score is lower than the threshold, the instance is considered negative (normal).  The AUC takes a value between 0.0 and 1.0 and, thus, can be expressed in percentage. 

In the context of anomaly detection and the evaluation of the performance of a model to differentiate between normal and abnormal sound data, the AUC is then given by \ref{auc}, where we are using the anomaly score of the normal data.
\begin{align}
AUC = \frac{1}{N_-N_+} \sum_{i=1}^{N_-}\sum_{j=1}^{N_+} \mathcal{H}\big( \mathcal{A} (x^{+}_j) - \mathcal{A} (x^{-}_i) \big)
\label{auc}
\end{align}

\begin{itemize}
\item $N_+$ and $N_-$: the number of normal and abnormal samples in test set, respectively.
\item $\{x^{-}_i \}_{i=1}^{N_-}$ and $\{x^{+}_j \}_{j=1}^{N_+}$ : the normal and abnormal test data vector, respectively, and sorted with their anomaly scores are in descending order.
\item $\mathcal{A} (x^{-}_i)$ : is the anomaly score of a normal test data $ x^{-}_i$.
\item $\mathcal{A} (x^{+}_j)$ : is the anomaly score of an abnormal test data $ x^{+}_j$.
\item 	
	$\mathcal{H}(x) =
	\begin{cases}
	1 & \text{if $x \geq \eta$} \\
	0 & \text{else}\\
	\end{cases}$
\item $\eta$ is a pre-specified threshold. 
\end{itemize}

\subsection{Baseline Model}

We compare the proposed approach with variational autoencoder (VAE) baseline. Our basic assumption is that this trained models will have high reconstruction error for anomalous machine sounds. The architecture of the neural networks are almost similar for VAE and HCVAE models. The main difference is that we concatenate the condition one-hot vector with input vector in HCVAE case, whereas in baseline VAE there is no condition. The neural network architecture can be summarized as follows: The encoder network comprises $FC(Input,128, ReLU)$; 3 $\times$ $F C(128,128, ReLU)$;  and $FC(128,8, ReLU)$, and the decoder network incorporates $FC(8,128, ReLU)$; 3 $\times$ $FC(128,128, ReLU)$;  and $FC(128, Output, none)$,  where $FC(a, b, f)$ means a fully connected layer with  $a$ input neurons,$b$ output neurons, and activation function $f$. The $ReLU$s are Rectified Linear Units \cite{jarrett2009best}. 


\section{Results and Discussion}
	Our main goal is to develop a method which can employ the available taxonomic knowledge into the model. Thus, we considered comparison between VAE and HCVAE in our study. As we discussed in the section I, we also checked domain adaptation capability and effects of inclusion of partial information into the model. To evaluate the anomaly detection performance, the AUC score is calculated for evaluation set. In Fig. \ref{fig_si}, we observe that HCVAE method (green bar) outperforms VAE method (orange bar) for almost all types of machines. In addition to this, to check the domain adaptation capability, we trained the VAE and HCVAE model with the weights which were pre-trained with development dataset. They are represented by VAE\_PT (dark blue bar) and HCVAE\_PT (light blue bar) in  the Fig. \ref{fig_si}. Finally, we also tested with only partial condition as input to HCVAE model. HCVAE\_Ci (gray bar) shows the results when we give only $C_i$ as additional input, and  HCVAE\_Cij (yellow bar) shows the results when we give only $C_{ij}$ as additional input. Note that HCVAE\_Ci and HCVAE\_Cij were also trained using pre-trained weights with development dataset. And HCVAE\_Ci is similar to conventional conditional variational autoencoder (CVAE).

	 From these results, we see that pre-trained weights are useful and improves the anomaly detection performance on new domain data. It is also interesting to see that partial conditional input also improves the accuracy. The reason could be that the real latent variable is distributed under $p_{\theta}(z|C)$. That is now a conditional probability distribution. This can be thought in a way that for each  possible value of $C$, we would have a $p(z)$ which helps us to represent the complicated distribution accurately and hence improves the anomaly detection performance.

\section{Conclusion}
 In this paper, we introduce hierarchical conditional autoencoder which provides better latent space representation. This model is utilized for anomaly detection task in machine sound. We achieved significant improvement in terms of AUC scores for all the different kind of machines with \textit{single} HCVAE model. This method shows that the generalization capability can be improved by the available taxonomic knowledge in the data. To check the generalization capabilities, we evaluated the model on different experimental settings. In the future, we will consider extending the current solution to a wider variety of sensors beyond microphones.

\bibliographystyle{IEEEtran}

\bibliography{refs}

\end{document}